\documentclass[letterpaper]{article}
\usepackage[margin=1in]{geometry}
\usepackage{graphicx,amsmath,amsthm,amssymb}
\usepackage{multirow}
\usepackage{natbib}
\usepackage{url,color,booktabs}
\usepackage{hyperref}

\def \ifempty#1{\def\temp{#1} \ifx\temp\empty }

\newcommand{\D}{\ensuremath{\mathbf{D}}}

\newcommand{\HH}{\ensuremath{\mathbf{H}}}
\newcommand{\I}{\ensuremath{\mathbf{I}}}

\newcommand{\M}{\ensuremath{\mathbf{M}}}

\newcommand{\Q}{\ensuremath{\mathbf{Q}}}

\newcommand{\U}{\ensuremath{\mathbf{U}}}
\newcommand{\V}{\ensuremath{\mathbf{V}}}

\newcommand{\X}{\ensuremath{\mathbf{X}}}

\newcommand{\z}{\ensuremath{\mathbf{z}}}

\newcommand{\bmu}{\ensuremath{\boldsymbol{\mu}}}

\newcommand{\bbR}{\ensuremath{\mathbb{R}}}

\newcommand{\calF}{\ensuremath{\mathcal{F}}}

\newcommand{\calT}{\ensuremath{\mathcal{T}}}

\newcommand{\norm}[2][]{%
  \ifempty{#1} {\left\lVert#2\right\rVert} \else {#1\lVert#2#1\rVert} \fi}

{%
\begin{list}{#1}{
\vspace{-\topsep}
\vspace{-\partopsep}
\setlength{\itemindent}{0cm}
\setlength{\rightmargin}{0cm}
\setlength{\listparindent}{0cm}
\settowidth{\labelwidth}{#1}
\setlength{\leftmargin}{\labelwidth}
\addtolength{\leftmargin}{\labelsep}
\setlength{\itemsep}{0cm}
}%
}%
{%
\end{list}
\vspace{-\topsep}
\vspace{-\partopsep}
}

{\begin{enumerate}%
}%
{\end{enumerate}}

\DeclareMathOperator*{\argmin}{arg\,min}
\DeclareMathOperator*{\argmax}{arg\,max}

\newcommand{\traceop}{\operatorname{tr}}
\newcommand{\trace}[1]{\ensuremath{\traceop\left(#1\right)}}

\graphicspath{{grf/}}

\bibpunct[, ]{(}{)}{;}{a}{,}{,} 
					
\title{Style Transfer by Rigid Alignment in Neural Net Feature Space}
\author{
   Suryabhan Singh Hada $\hspace{2ex}$ Miguel {\'A}.\ Carreira-Perpi{\~n}{\'a}n\\
   Dept. CSE, University of California, Merced\\
   {\url{http://eecs.ucmerced.edu}}
	}
\date{December 23, 2020}

\definecolor{DarkRed}{rgb}{0.545,0,0}
\hypersetup{
  breaklinks=true, colorlinks=true, linkcolor=black,
  citecolor=black, urlcolor=DarkRed 
}

\begin{document}
					
\maketitle
\begin{abstract}
 Arbitrary style transfer is an important problem in computer vision that aims to transfer style patterns from an arbitrary style image to a given content image. However, current methods either rely on slow iterative optimization or fast pre-determined feature transformation, but at the cost of compromised visual quality of the styled image; especially, distorted content structure. In this work, we present an effective and efficient approach for arbitrary style transfer that seamlessly transfers style patterns as well as keep content structure intact in the styled image. We achieve this by aligning style features to content features using rigid alignment; thus modifying style features, unlike the existing methods that do the opposite. We demonstrate the effectiveness of the proposed approach by generating high-quality stylized images and compare the results with the current state-of-the-art techniques for arbitrary style transfer.
\end{abstract}

\section{Introduction}

Given a pair of style and a target image, style transfer is a process of transferring the texture of the style image to the target image, keeping the structure of the target image unchanged. 
\emph{Most of the recent work in the neural style transfer is based on the implicit hypothesis is that working in deep neural network feature space can transfer texture and other high-level information from one image to another without altering the image structure much.} Recent work from \citet{Gatys_16a} (Neural style transfer (NST)) shows the power of the Convolution Neural Networks (CNN) in style transfer. 

In just a few years, significant effort has been made to improve NST, either by iterative optimization-based approaches \citep{LiWand16a, Li_17a, Risser_17a} or feed-forward network approximation \citep{Johnson_16a, Ulyanov_16b,  Ulyanov_16a, LiWand16b, Dumoul_17a, Chen_17a, Li_17f, Shen_18a, ZhangDana17a, Wang_17c}. Optimization-based methods \citep{Gatys_16a, LiWand16a, Li_17a, Risser_17a},  achieve visually great results, but at the cost of efficiency, as every style transfer requires multiple optimization steps.  On the other hand,  feed-forward network-based style transfer methods \citep{Johnson_16a, Ulyanov_16b,  Ulyanov_16a, LiWand16b, Dumoul_17a, Chen_17a, Li_17f, Shen_18a, ZhangDana17a, Wang_17c} provide efficiency and quality, but at the cost of generalization. These networks are limited to a fixed number of styles. 

Arbitrary style transfer can achieve generalization, quality, and efficiency at the same time. The goal is to find a transformation that can take style and content features as input, and produce a stylized feature that does not compromise reconstructed stylized image quality. 

However, current works in this regard \citep{HuangBelong17a, Li_17d, ChenSchmidt16a, Sheng_18a} have failed in the quality of the generated results.  Among these
\citet{HuangBelong17a} and \citet{ChenSchmidt16a} use external style signals to supervise the content modification on a feed-forward network.  The network is trained by using perpetual loss \citep{Johnson_16a},  which is known to be unstable and produce unsatisfactory style transfer results \citep{Gupta_17b, Risser_17a}.

On the contrary,  \citet{Li_17d},  \citet{ChenSchmidt16a} and \citet{Sheng_18a} manipulate the content features under the guidance of the style features in a shared high-level feature space.  By decoding the manipulated features back into the image space with a style-agnostic image decoder,  the reconstructed images will be stylized with seamless integration of the style patterns.  However,  these techniques over-distort the content or fail to balance the low level and global style patterns. 
%

In this work, we address the aforementioned issues by modifying style features instead of content features during style transfer.  \emph{Our hypothesis is if we consider images as a collection of points in feature space, where each point represents some spatial information, and if we align these points clouds using rigid alignment, we can transform these points without introducing any distortion.} By doing so, we solve the problem of content over-distortion since alignment does not manipulate content features.  Similar to \citet{Li_17d} and \citet{Sheng_18a},  our method does not require any training and can be applied to any style image in real-time.  We also provide comprehensive evaluations to compare with the prior arbitrary style transfer methods \citep{Gatys_16a, HuangBelong17a, Li_17d, Sheng_18a}, to show that our method achieves state-of-the-art performance.

Our contributions in this paper are threefold: 1) We achieve style transfer by using rigid alignment, which is different from traditional style transfer methods that depend on feature statistics matching. Rigid alignment is well studied in computer vision for many years and has been very successful in image registration and many problems of that type. We show that by rearranging the content and style features in a specific manner (each channel ($C$) as a point in $\bbR^{HW}$ space, where $H$ is height, and $W$ is the width of the feature), they can be considered as a point cloud of $C$  points. 2) We provide a closed-form solution to the style transfer problem. 3) The proposed approach achieves impressive style transfer results in real-time without introducing content distortion. 

\begin{figure*}[t]
 \centering
\begin{tabular}{@{}c@{}}
{\hspace{0.00\linewidth}content \hspace{0.1\linewidth} style \hspace{0.1\linewidth}Avatar[\citenum{Sheng_18a}]  \hspace{0.07\linewidth} WCT[\citenum{Li_17d}]\hspace{0.07\linewidth} AdaIN[\citenum{HuangBelong17a}] \hspace{0.075\linewidth}Ours \hspace{0.0\linewidth}}\\
\includegraphics*[width=1\linewidth]{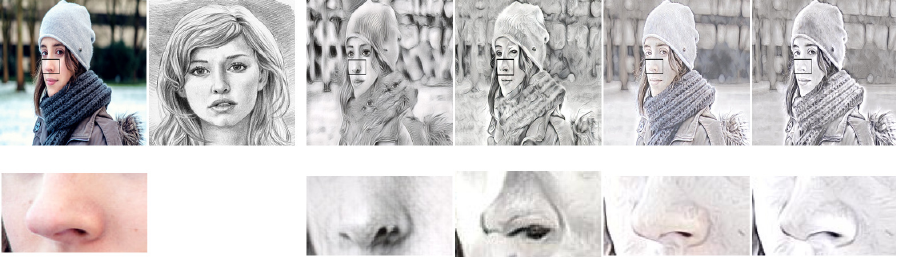}
\end{tabular}
\caption{Content distortion during style transfer. Regions marked by bounding boxes are zoomed in for a better visualization.}
\label{f:closeup}
\end{figure*}

\section{Related work}
Due to the wide variety of applications, the problem of style transfer has been studied for a long time in computer vision. Before seminal work by~\citet{Gatys_16a}, the problem of style transfer has been focused as \emph{non-photorealistic rendering (NPR)} \citep{Kyprian_12a}, and closely related to texture synthesis \citep{EfrosFreeman01a,EfrosLeung99a}. Early approaches rely on finding low-level image correspondence and do not capture high-level semantic information well. As mentioned above, the use of CNN features in style transfer has improved the results significantly. We can divide the current Neural style transfer literature into four parts.

\begin{itemize}

\item \textbf{Slow optimization-based methods:} \citet{Gatys_16a} introduced the first  NST method for style transfer. The authors created artistic style transfer by matching multi-level feature statistics of content and style images extracted from a pre-trained image classification CNN (VGG \citep{SimonyZisser15a}) using Gram matrix. Soon after this, other variations were introduced to achieve better style transfer \citep{LiWand16a,Li_17a,Risser_17a}, user controls like spatial control and color preserving \citep{Gatys_16b, Risser_17a} or include semantic information \citep{Frigo_16a, Champand16a}. However, these methods require an iterative optimization over the image, which makes it impossible to apply in real-time.

\item \textbf{Single style feed-forward networks:}  Recently, \citet{Johnson_16a}, \citet{Ulyanov_16b}, \citet{Ulyanov_16a} and \citet{LiWand16b} address the real-time issue by approximating the iterative back-propagating procedure to feed-forward neural networks, trained either by the perceptual loss \citep{Johnson_16a, Ulyanov_16b} or Markovian generative adversarial loss \citep{LiWand16b}. Although these approaches achieve style transfer in real-time, they require training a new model for every style. This makes them very difficult to use for multiple styles, as every single style requires hours of training.

\item \textbf{Single network for multiple styles:} Later \citet{Dumoul_17a}, \citet{Chen_17a}, \citet{Li_17f} and \citet{Shen_18a} have tried to tackle the problem of multiple styles by training a small number of parameters for every new style while keeping rest of the network the same. \emph{Conditional instance normalization} \citep{Dumoul_17a} achieved it by training channel-wise statistics corresponding to each style. Stylebank \citep{Chen_17a} learned convolution filters for each style, \citet{Li_17f} transferred styles by binary selection units and \citet{Shen_18a} trained a meta-network that generates a $14$ layer network for each content and style image pair. On the other hand, \citet{ZhangDana17a} trained a weight matrix to combine style and content features. The major drawback is the model size that grows proportionally to the number of style images. Additionally, there is interference among different styles \citep{Jing_17a}, which affects stylization quality.

\item  \textbf{Single network for arbitrary styles:} Some recent works \citep{HuangBelong17a,Li_17d,ChenSchmidt16a,Sheng_18a, Gu_18c} have been focused on creating a single model for arbitrary style i.e., one model for any style. \citet{Gu_18c} rearrange style features patches with respect to content features patches. However, this requires solving an optimization problem to find the nearest neighbor, which is slow, thus not suitable for real-time use. 
\citet{ChenSchmidt16a} swaps the content feature patches with the closest style feature patch but fails if the domain gap between content and style is large.  \citet{Sheng_18a} addresses this problem by first normalizing the features and then apply the patch swapping. Although this improves the stylization quality, it still produces content distortion and misses global style patterns, as shown in fig.~\ref{f:closeup}.  WCT \citep{Li_17d} transfers multi-level style patterns by recursively applying whitening and coloring transformation (WCT) to a set of trained auto-encoders with different levels. However, similar to \citet{Sheng_18a}, WCT also produces content distortion; moreover, this introduces some unwanted patterns in the styled image \citep{Jing_17a}.
Adaptive Instance normalization (AdaIN) \citep{HuangBelong17a} matches the channel-wise statistics (mean and variance) of content features to the style features, but this matching occurs only at one layer, which authors try to compensate by training a network on perpetual loss \citep{Johnson_16a}. Although this does not introduce content distortion, it fails to capture style patterns.  
\end{itemize}

\begin{figure*}[t]
 \centering
 \begin{tabular}{@{\hspace{8ex}}c@{\hspace{17ex}}c@{\hspace{14ex}} c@{\hspace{5ex}}c@{}}
\hspace{3ex}{content}&{style}&\hspace{8ex}{relu 1 to 4 }&{ relu\_4}\\
\multicolumn{4}{c}{\includegraphics*[width=0.95\linewidth]{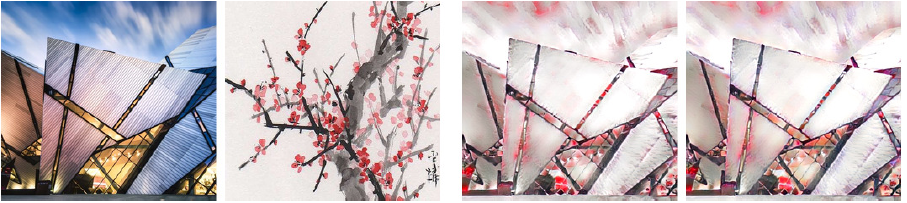}}
\end{tabular}
\caption{Comparison between the style transfer results, by applying rigid alignment only at the deepest layer (relu\_4) instead of every layer. The third image shows the style transfer result by applying alignment at every layer (\{relu\_1, relu\_2, relu\_3, relu\_4\}). On the other hand, the last column shows the style transfer result by applying alignment only at the deepest layer (relu\_4). Both produce nearly identical results.}
\label{f:proc_once}
\end{figure*}

\begin{figure*}[t]
 \centering
  \begin{tabular}{@{}c@{}c@{\hspace{2ex}}c@{}c@{\hspace{.3ex}}c@{}c@{\hspace{.5ex}}c@{}}
  
&&\rotatebox{90}{\hspace{.8ex}{frames}}&&
\includegraphics*[width=0.8\linewidth]{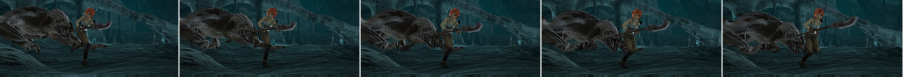}\\ \\

{\vspace{2.5ex}Style image}&&
\rotatebox{90}{\hspace{.8ex}{Ours}}&&
\includegraphics*[width=0.8\linewidth]{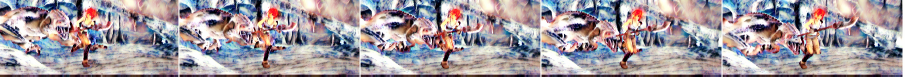}\\

\multirow{2}{*}[ 8ex]{\includegraphics*[width=0.16\linewidth]{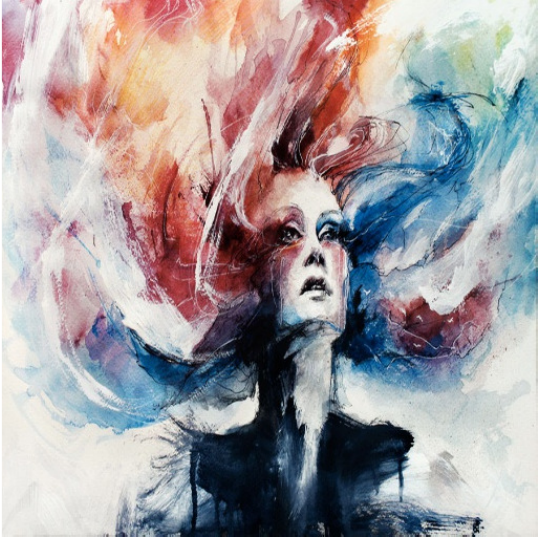}}
&&\rotatebox{90}{\hspace{.8ex}{WCT}}&&
\includegraphics*[width=0.8\linewidth]{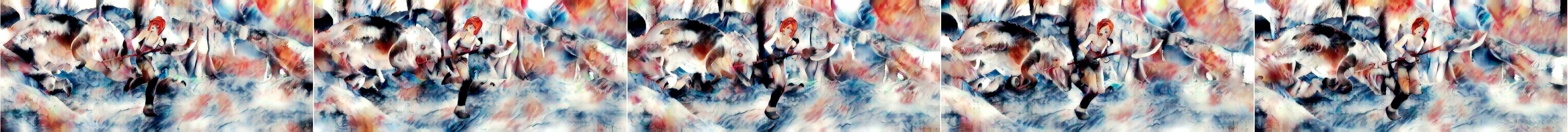}\\

&&\rotatebox{90}{\hspace{.3ex}{Avatar}}&&
\includegraphics*[width=0.8\linewidth]{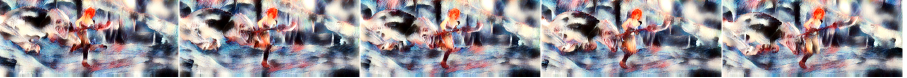}

\end{tabular}

\caption{Video stylization using the proposed approach. Similar to WCT~\citep{Li_17d} and Avatar-Net~\citep{Sheng_18a}, the proposed method keeps style patterns coherent in each frame. However, unlike the other two, the proposed method does not suffer from content distortion. In the case of WCT~\citep{Li_17d}, the distortion is much worse than Avatar-Net, especially the animal's face. Animations are provided at author’s webpage.}
\label{f:video}
\end{figure*}  

The common part of the existing arbitrary style transfer methods, that they all try to modify the content features during the style transfer process. This eventually creates content distortion. Different from existing methods, our approach manipulates the style features during style transfer. We achieve this in two steps. First, we apply channel-wise moment matching (mean and variance) between content and style features, just as AdaIN \citep{ HuangBelong17a}. Second, we use rigid alignment (Procrustes analysis~\citep[see][chap.~21]{BorgGroenen97a}) to align style features to content features. This alignment modifies the style features to adapt content structure, thus avoiding any content distortions while keeping its style information intact. In the next sections, we describe our complete approach.



\section{Style transfer in neural net features space}
Generally Speaking style transfer as follows
Let $\z_c \in \bbR^{C \times H \times W}$ is a feature extracted from a layer of a pre-trained CNN when the content image passes through the network. Here, $H$ is the height, $W$ is the width, and $C$ is the number of channels of the feature $\z_c$. Similarly, for style image $\z_s \in \bbR^{C \times H \times W}$ represents the corresponding features.

For any arbitrary style transfer method, we pass $\z_s$ and $\z_c$ to a transformation function $\calT$  which outputs styled feature $\z_{cs}$ as described in eq.~\eqref{e:gen_styled}:
\begin{equation}
\z_{cs} = \calT(\z_c, \z_s).
\label{e:gen_styled}
\end{equation}  
Reconstruction of $\z_{cs}$ to image space gives the styled image. The difficult part is finding the transformation function $\calT$ that is style-agnostic like \citet{Sheng_18a},\citet{ChenSchmidt16a} and \citet{Li_17d}, but unlike these, it captures local and global style information without distorting the content and does not need iterative optimization.
\section{Proposed approach }
\label{s:procrustes}

Although AdaIN \citep{HuangBelong17a} is not style agnostic, it involves a transformation which is entirely style agnostic: channel-wise moment matching.
This involves matching channel-wise mean and variance of content features to those of style features as follows: 
 \begin{equation}
\z_{c^{\prime}}= \bigg( \frac{\z_c - \calF_{\mu}(\z_c)}{\calF_{\sigma}(\z_c)} \bigg) \calF_{\sigma}(\z_s) + \calF_{\mu}(\z_s).
\label{e:adaIN}
\end{equation}
Here,  $\calF_{\mu}(.)$ and  $ \calF_{\sigma}(.)$ is channel-wise mean and variance respectively. Although this channel-wise alignment produces unsatisfactory styled results, it is able to transfer local patterns of style image without distorting content structure as shown in fig.~\ref{f:closeup}. 
Moment matching does not provide a perfect alignment among channels of style and content features which leads to missing global style patterns and thus unsatisfactory styled results. Other approaches  achieve this, either by doing WCT transformation \citep{Li_17d}  or patch replacement \citep{Sheng_18a,ChenSchmidt16a}, but this requires content features modification  that leads to content distortion. We tackle this, by aligning style features to content features instead. In that way, style features get structure of content while maintaining their global patterns.  

One simple way of alignment that prevents distortion is rigid alignment~\citep{ BorgGroenen97a} and (scaling). This involves shifting, scaling and finally rotation of the points that to be moved (styled features) with respect to the target points (content features after moment matching). For this we consider both features as point clouds of size $C$ with each point is in $ \bbR^{HW}$ space, i.e. $\z_c, \z_s \in \bbR ^{ C \times HW}$.  Now, we apply rigid transformation in following steps:
\begin{itemize}
\item \textbf{Step-I: Shifting.} First, we need to shift both point clouds $\z_c$ and $\z_s$ to a common point in $\bbR^{HW}$ space. We center these point clouds to the origin as follows:
\begin{align}
\bar{\z}_c = \z_c -\bmu_c \nonumber \\
\bar{\z}_s = \z_s -\bmu_s . 
\end{align}
Here, $\bmu_c$ and $\bmu_s \in \bbR^{HW}$ are the mean of the $\z_c$ and $\z_s$ point clouds respectively.

\item \textbf{Step-II: Scaling.} Both point clouds need to have the same scale before alignment. For this, we make each point cloud to have unit Frobenius norm:
\begin{align}
\hat{\z}_c =  \frac{\bar{\z}_c}{\norm{\z_c}_F} \nonumber \\
\hat{\z}_s =  \frac{\bar{\z}_s}{\norm{\z_s}_F} .
\end{align}
Here, $\norm{.}_F$ represents Frobenius norm.

\item \textbf{Step-III: Rotation.} Next step involves rotation of $\hat{\z}_s$ so that it can align perfectly with $\hat{\z}_c$. For this, we multiply $\hat{\z}_s$ to a rotation matrix that can be created as follows:  
\begin{align}
 \argmin_\Q \norm{\hat{\z}_s \Q -\hat{\z}_c}_2^2 \quad \text{s.t.} \quad \Q\text{ is orthogonal} .
 \label{e:roation}
\end{align}
Although this is an optimization problem, it can be solved as follows: 
\begin{align}
\norm{\hat{\z}_s \Q -\hat{\z}_c}_2^2 = \trace{\hat{\z}_s^T \hat{\z}_s +\hat{\z}_c^T \hat{\z}_c } -2 \trace {\hat{\z}_c^T \hat{\z}_s \Q} .
\end{align}
Since, $\trace{\hat{\z}_s^T \hat{\z}_s +\hat{\z}_c^T \hat{\z}_c }$ term is independent of $\Q$, so eq.~\eqref{e:roation} becomes:

\begin{align}
 \argmax_\Q \trace{\hat{\z}_c^T \hat{\z}_s\Q} \quad \text{s.t.} \quad \Q\text{ is orthogonal} .
\end{align}
Using singular value decomposition of $\hat{\z}_c^T \hat{\z}_s= \U \mathbf{S} \V^T$ and cyclic property of trace we have:

\begin{align}
\trace{\hat{\z}_c^T \hat{\z}_s\Q} &= \trace{\U \mathbf{S} \V^T \Q} \nonumber \\
& =  \trace{\mathbf{S} \V^T \Q \U} \nonumber  \\
&= \trace{\mathbf{S} \HH} .
\label{e:svd_part}
\end{align}
Here, $\HH= \V^T \Q \U$ is an orthogonal matrix, as it is product of orthogonal matrices. Since, $\mathbf{S}$ is a diagonal matrix, so in order to maximize  $\trace{\mathbf{S} \HH}$, the diagonal values of $\HH$ need to equal to  $1$. Now, we have: 

\begin{align}
\HH  = \V^T\Q\U &= \I \nonumber \\
\text{or ,} \qquad \Q &= \V \U^T .
\end{align}

\item \textbf{Step-IV: Alignment.} After obtaining rotation matrix $\Q$, we scale and shift style point cloud with respect to the original content features in the following way:
\begin{align}
\z_{sc} = \norm{\z_c}_F \hat{\z}_s \Q + \bmu_c
\label{e:proc}
\end{align} 
$\z_{sc}$ is the final styled feature.

This alignment makes style features to adapt content structure while keeping its local and global patterns intact. 

\textbf{Note:} Above we assume that both $\z_c$ and $\z_s$ are of equal size, so as to make the explanation easy. In case of $\z_c\in \bbR^{C \times H_c W_c}$ and  $\z_s\in \bbR^{C \times H_s W_s}$, the only change will be in eq.~\eqref{e:roation} where the orthogonal matrix $\Q$ is rectangular and satisfies $\Q^T \Q = \I$ (i.e. $\Q \in \bbR^{H_s W_s \times H_c W_c}$).

\end{itemize}

\begin{figure*}[t!]
 \centering
 \begin{tabular}{@{}c@{}c@{}c@{\hspace{2ex}}c@{}}
 \vspace{2ex}
 \rotatebox{90}{\hspace{1.2ex}{content}}&
 \includegraphics*[width=0.15\linewidth]{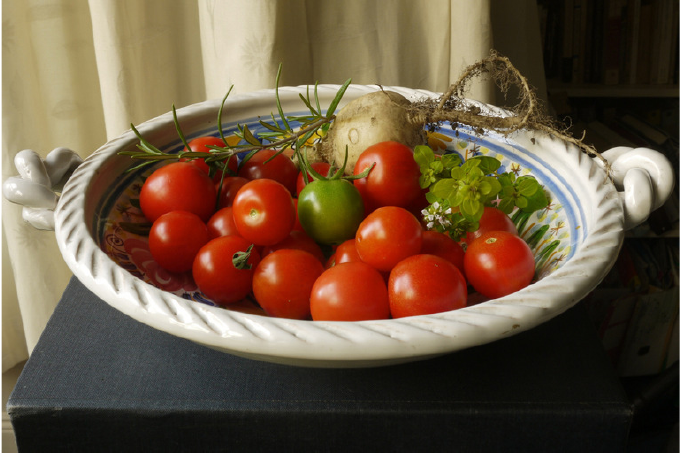}&&
 \multirow{2}{*}[10ex]{\includegraphics*[width=0.75\linewidth]{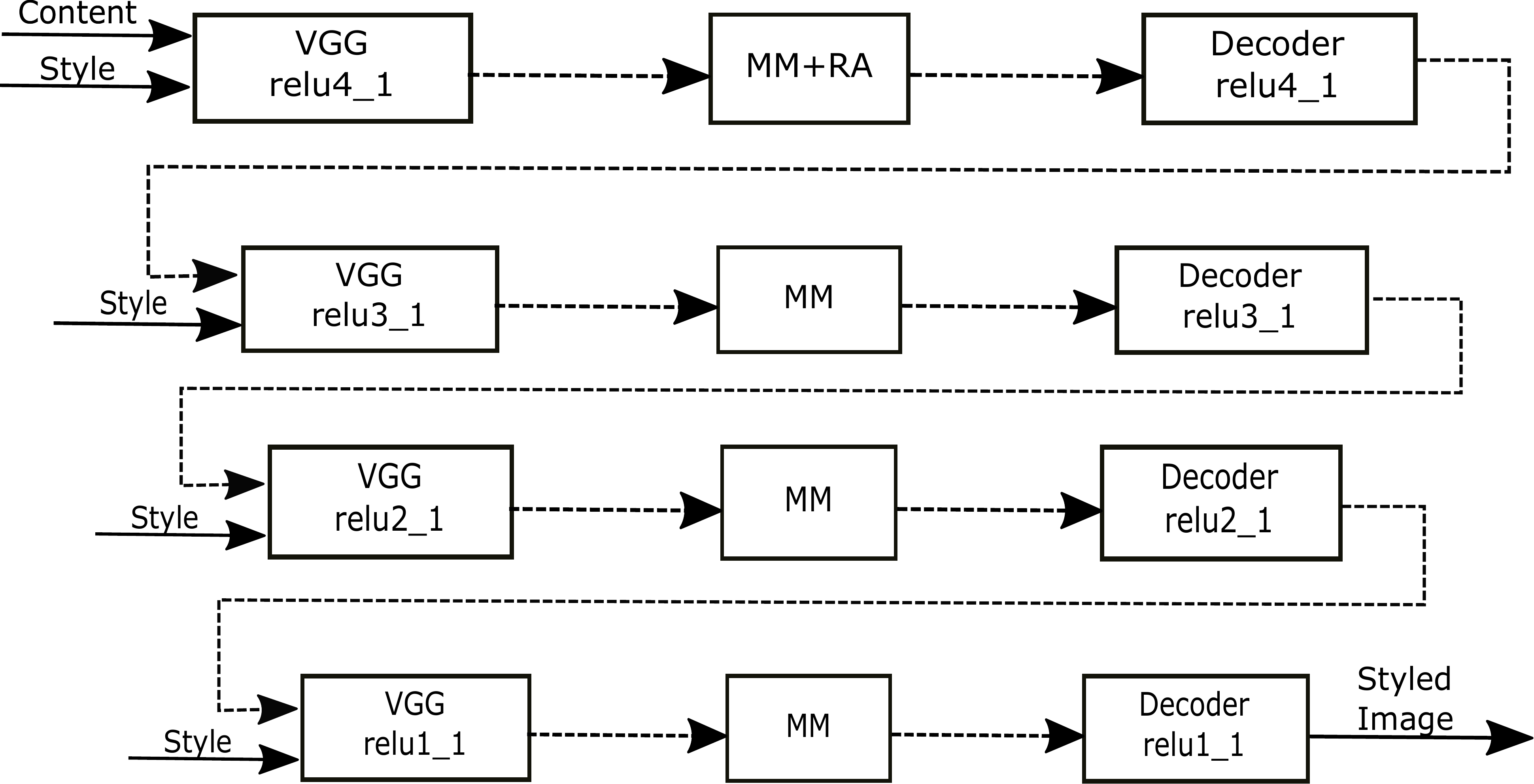}}\\
 \rotatebox{90}{\hspace{4ex}{style}}&
 \includegraphics*[width=0.15\linewidth]{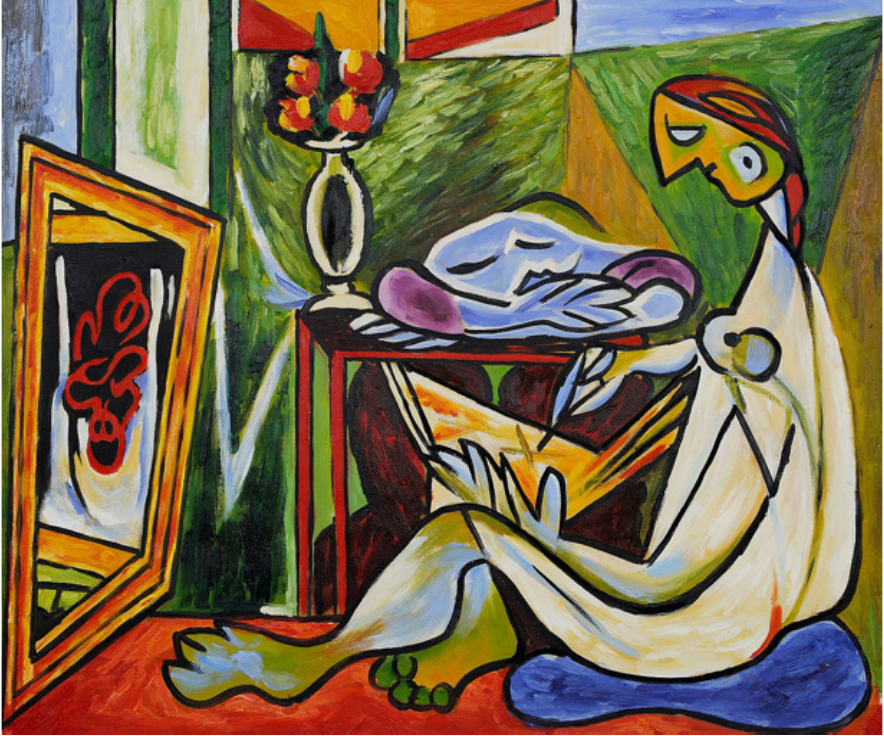}&&
\end{tabular} 
 
\vspace{16ex}
 
 \begin{tabular}{@{}c@{}}
\includegraphics*[width=1\linewidth]{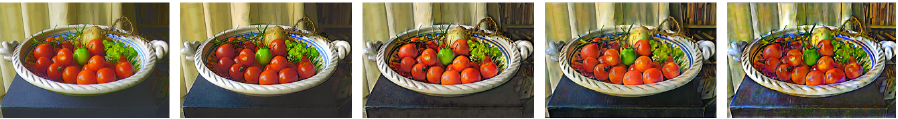}\\
{\hspace{2ex}relu\_1 \hspace{15ex}relu\_2 \hspace{15ex}relu\_3 \hspace{15ex}relu\_4 \hspace{13ex}relu 1 to 4}\\
\end{tabular} 
\caption{\emph{Top:} Network pipeline of the proposed style transfer method that is similar to \citet{Li_17d}. The result obtained by matching higher-level statistics of the style is treated as the new content to continue to match lower-level information of the style. MM represents moment matching and RA represents rigid alignment. \emph{Bottom:} Comparison between single-level and multi-level stylization with the proposed approach. The first four images show styled images created by applying moment matching and rigid alignment to individual VGG features. The last image shows stylization results by applying multi-level stylization, as shown in the above network pipeline.}
\label{f:multi_lvl}
\end{figure*}

\subsection{Multi-level style transfer}
\label{s:mutlistyle}
As shown in the \citet{Gatys_16a}, features from different layers provide different details during style transfer. Lower layer features (\emph{relu\_1} and \emph{relu\_2}) provide color and texture information, while features from higher layer (\emph{relu\_3} and \emph{relu\_4}) provide common patterns details (fig.~\ref{f:multi_lvl}). Similar to WCT \citep{Li_17d}, we also do this by cascading the image through different auto-encoders. However, unlike WCT \citep{Li_17d}  we do not need to do the alignment described in section~\ref{s:procrustes} at every level. We only apply the alignment at the deepest layer (\emph{relu4\_1}).

Doing alignment at each layer or only at deepest layer (\emph{relu4\_1}) produce identical results as shown in  fig.~\ref{f:proc_once}. This also shows the rigid alignment of style features to content is perfect. 

Once the features are aligned, we only need to take care of local textures at other layers. We do this by applying moment matching (eq.~\eqref{e:adaIN}) at lower layers.  The complete pipeline is shown in figure:~\ref{f:multi_lvl}.

\section{Need to arrange features in $\bbR^{C\times HW}$ space}
\begin{figure*}[t!]
 \centering
 \begin{tabular}{@{\hspace{8ex}}c@{\hspace{17ex}}c@{\hspace{14ex}} c@{\hspace{5ex}}c@{}}
\hspace{5ex}{content}&{style}&\hspace{10ex}{$C\times HW$}&{$HW\times C$}\\
\multicolumn{4}{c}{\includegraphics*[width=0.98\linewidth]{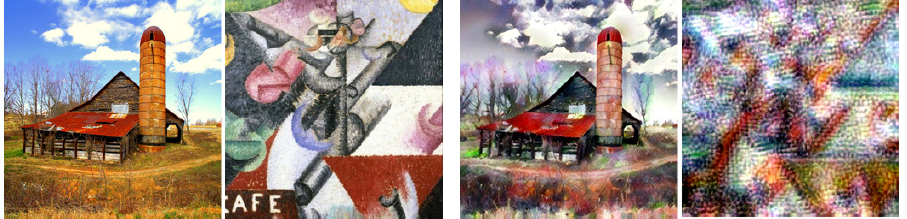}}
\end{tabular}
\caption{\emph{Third column:} style transfer with features ($\z$) transformed as $C$ cloud points and each in $\bbR^{HW}$ space. \emph{Fourth column:} style transfer with $HW$ cloud points and each in $\bbR^{C}$ space.}
\label{f:configs}
\end{figure*}

As mentioned above, for alignment we consider the deep neural network features ($\z \in \bbR^{C\times H \times W}$) as a point cloud which has $C$ points each of dimension $HW$. We can also choose another configuration where each point is in $\bbR^C$  space, thus having $HW$ points in the point cloud. In fig.~\ref{f:configs}, we show a comparison of style transfer with the two configurations. As shown in the fig.~\ref{f:configs}, having the later configuration results in complete distortion of content structure in the final styled image. The reason for that is deep neural network features (convolution layers) preserve some spatial structure, which is required for style transfer and successful image reconstruction. Therefore, we need to transform the features in a specific manner so that we do not lose the spatial structure after alignment. That is why, for alignment, we transform $\z$ such that the point cloud has $C$ points each of dimension $HW$.

\section{Experiments }
\subsection{Decoder training}
We use a pre-trained auto-encoder network from \citet{Li_17d}. This auto-encoder network has been trained for general image reconstruction. The encoder part of the network is the pre-trained VGG-19 \citep{SimonyZisser15a} that has been fixed, and the decoder network ($\D$) is trained to invert the VGG features to image space. As mentioned in \citet{Li_17d}, the decoder is designed as being symmetrical to that of the VGG-19 network, with the nearest neighbor up-sampling layer used as the inverse of max pool layers. 
 \citet{Li_17d} trained five decoders for reconstructing images from features extracted at different layers of the VGG-19 network. These layers are \emph{relu5\_1,  relu4\_1, relu3\_1, relu2\_1,} and  \emph{relu1\_1}. The loss function for training involves pixel reconstruction loss and feature loss \citep{DosovitBrox16b}:
\begin{align}
\argmin_\theta \norm{\X- \D_\theta(\z)}^2_2 + \lambda \norm{\Phi_l(\X)- \Phi_l(\D_\theta(\z))}^2_2
\end{align} 
where $\theta$ are the weights of the decoder $\D$. $\X$, $\z$ are the original image and corresponding VGG features, respectively, and $\Phi_l(\X)$ is a VGG-19 encoder that extracts features from layer $l$. In addition, $\lambda$ is the weight to balance the two losses. The decoders have been trained on the Microsoft COCO dataset \citep{Lin_14d}. However, unlike \citet{Li_17d}, we use only four decoders in our experiments for multi-level style transfer. These decoders correspond to \emph{ relu4\_1, relu3\_1, relu2\_1,} and \emph{relu1\_1} layers of the VGG-19 network.

\begin{figure*}[t!]
 \centering
 \begin{tabular}{@{\hspace{5ex}}c@{\hspace{9ex}}  c@{\hspace{5ex}}c@{\hspace{5ex}} c@{\hspace{5ex}}  c@{\hspace{5ex}} c@{\hspace{4ex}}  c@{\hspace{2ex}}  c@{}}
 {content}& {style} && {Gatys[\citenum{Gatys_16a}] }& {AdaIN [\citenum{HuangBelong17a}]} & {WCT [\citenum{Li_17d}]} & {Avatar [\citenum{Sheng_18a}]}& {Ours}\\
\multicolumn{8}{l}{\includegraphics*[width = 1\linewidth]{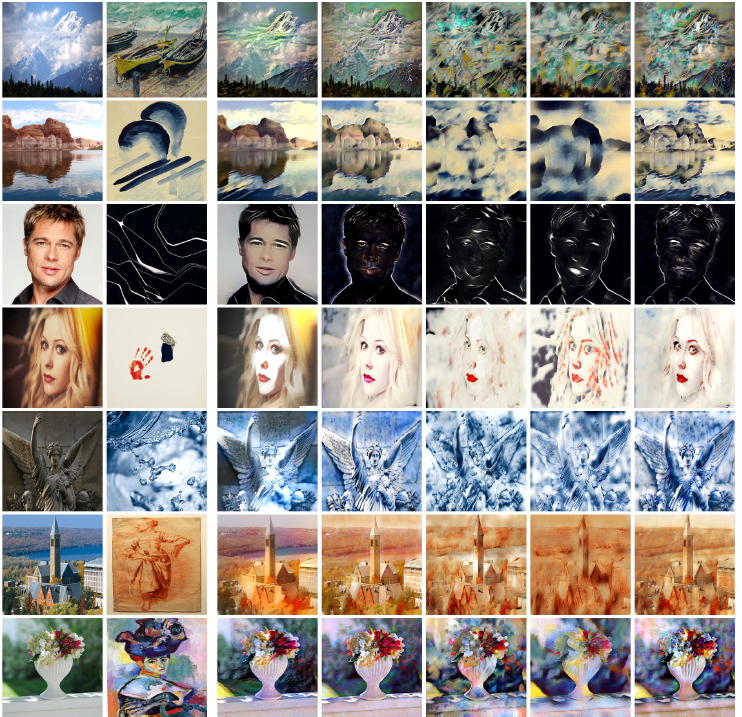}}
\end{tabular}
\caption{Figure shows comparison of our style transfer approach with existing work. }
\label{f:comaprison}
\end{figure*}

\subsection{Comparison with prior style transfer methods}
To show the effectiveness of the proposed method, we compare our results with two types of arbitrary style transfer approaches. The first type is iterative optimization-based \citep{Gatys_16a} and the second type is fast arbitrary style transfer method \citep{Li_17d, Shen_18a, HuangBelong17a}. We present these stylization results in fig.~\ref{f:comaprison}. 

Although optimization-based approach \citep{Gatys_16a} performs arbitrary style transfer, it requires slow optimization for this. Moreover, it suffers from getting stuck at a bad local minimum. This results in visually unsatisfied style transfer results, as shown in the third and fourth rows.  AdaIN \citep{HuangBelong17a} addresses the issue of local minima along with efficiency but fails to capture the style patterns. For instance, in the third row, the styled image contains colors from the content, such as red color on the lips. Contrary to this, WCT \citep{Li_17d} and Avatar-Net \citep{Shen_18a} perform very well in capturing the style patterns by matching second-order statistics and the latter one by normalized patch swapping. However, both methods fail to maintain the content structure in the stylized results.  For instance, in the first row, WCT \citep{Li_17d} completely destroys the content structure: mountains and clouds are indistinguishable. Similarly, in the second and fifth row, content image details are too distorted. Although Avatar-Net \citep{Shen_18a} performs better than WCT \citep{Li_17d} as in the first and fifth rows, it fails too in maintaining content information, as shown in the second and sixth rows. In the second row, the styled image does not even have any content information.

On the other hand, the proposed method not only captures style patterns similar to WCT \citep{Li_17d} and Avatar-Net \citep{Shen_18a} but also maintains the content structure perfectly as shown in the first, second, and fifth row where the other two failed.

We also provide a close-up in fig.~\ref{f:closeup}. As shown in the figure, WCT \citep{Li_17d} and Avatar-Net \citep{Shen_18a} distort the content image structure. The nose in the styled image is too much distorted, making these methods difficult to use with human faces. Contrary to this, AdaIN \citep{HuangBelong17a} and the proposed method keep content information intact, as shown in the last two columns of the second row. However, AdaIN \citep{HuangBelong17a} does not capture style patterns very well. On the other hand, the proposed method captures style patterns very well without any content distortion in the styled image.

In addition to image-based stylization,  the proposed method can also do video stylization. We achieve this by just doing per-frame style transfer, as shown in fig.~\ref{f:video}. The styled video is coherent over adjacent frames since the style features adjust themselves instead of content, so the style transfer is spatially invariant and robust to small content variations. In contrast, Avatar-Net \citep{Sheng_18a} and WCT \citep{Li_17d} contain severe content distortions, with the distortion is much worse in WCT \citep{Li_17d}.

\subsection{Efficiency}
We compare the execution time for style transfer of the proposed method with state-of-the-art arbitrary style transfer methods in the table~\ref{t:exec_time}. We implement all methods in Tensorflow~\citep{Abadi_16a} for a fair comparison.~\citet{Gatys_16a} approach is very slow due to iterative optimization steps that involve multiple forward and backward pass through a pre-trained network. On the contrary, other methods have very good execution time, as these methods are feed-forward network based. Among all, AdaIN \citep{HuangBelong17a} performs best since it requires only moment-matching between content and style features. WCT \citep{Li_17d} is relatively slower as it requires SVD operation at each layer during multi-layer style transfer. Avatar-Net \citep{Sheng_18a} has better execution time compared to WCT \citep{Li_17d} and ours. This is because of the GPU based style-swap layer and hour-glass multi-layer network. 

On the other hand, our method is comparatively slower than AdaIN \citep{HuangBelong17a}, and Avatar-Net \citep{Sheng_18a} as our method involves SVD operation at \emph{relu\_4}. Additionally, it requires to pass through multiple auto-encoders for multi-level style transfer similar to WCT \citep{Li_17d}. However, unlike WCT \citep{Li_17d} proposed method needs only one SVD operation as shown in fig.~\ref{f:proc_once} and thus have better execution time compared to WCT \citep{Li_17d}.
  
\begin{table}[t!]
\centering 
\begin{tabular}{@{}c |c@{\hspace{2ex}}c@{}}
\toprule
{\textbf{Method}}&&{\textbf{Execution time (in sec) ($512 \times 512$)}} \\
\midrule
{Gatys~\citep{Gatys_16a}} && 58 \\
{AdaIN~\citep{HuangBelong17a}} && 0.13 \\
{WCT~\citep{Li_17d}} && 1.12 \\
{Avatar-Net~\citep{Sheng_18a}} && 0.34 \\
{Ours} && 0.46 \\
\bottomrule
\end{tabular}
\caption{Execution time (in seconds) comparison for style transfer among the proposed method and state of the art methods. }
\label{t:exec_time}
\end{table}

\subsection{Numeric comparison}
In table~\ref{t:numeric_comp} we show numerical comparison between different style methods. We provide average content loss ($L_c$) and style loss ($L_s$) from \citet{Gatys_16a}, for the images in fig.~\ref{f:comaprison}:

\begin{align}
L_c &= \frac{1}{2CHW} \sum_{i,j} \norm{\z_{c_{i,j}}-\z_{i,j}}^2_2 \\
L_s &= \frac{1}{4C^2H^2W^2} \sum_{i,j}\norm{G_{i,j}(\z_s)-G_{i,j}(\z)}^2_2 .
\end{align}
Here, $\z_c$ is the content feature, $\z_s$ is the style feature, $\z$ is the styled feature, and $G(.)$ provides the Gram matrix. As shown in the table~\ref{t:numeric_comp}, WCT~\citep{Li_17d} and Avatar-Net~\citep{Sheng_18a} have smaller style losses because these methods prefer more style patterns in the styled result. However, as shown in fig.~\ref{f:closeup} and~\ref{f:comaprison} this leads to content distortion. On the other hand, AdaIN~\citep{HuangBelong17a} performs better in terms of content loss as it maintains more content information, but this produces results with fewer style patterns. So, any method that performs best in either content loss or style loss will produce unsatisfactory styled results. A good style transfer method should perform somewhere in between, which the proposed method achieves. The proposed method not only performs well in terms of content loss but is also on par with WCT \citep{Li_17d} and Avatar-Net \citep{Sheng_18a} in terms of style loss. This proves our intuition that by aligning style features to content features, we not only preserve content structure but also effectively transfers style patterns.

\begin{table}[t!]
\centering 
\begin{tabular}{c|c@{\hspace{5ex}}c@{\hspace{5ex}}|@{\hspace{5ex}}c@{\hspace{5ex}}}
\toprule
{\textbf{\hspace{4ex}\textbf{Method}\hspace{4ex}}}&&$\log(L_c)$&$\log(L_s)$\\
\midrule
{Gatys~\citep{Gatys_16a}} && \textbf{4.40}&8.28\\
{AdaIN~\citep{HuangBelong17a}} && 4.62&8.18\\
{WCT~\citep{Li_17d}}&& 4.79&7.83\\
{Avatar-Net~\citep{Sheng_18a}} && 4.75&\textbf{7.77}\\
\hline
{Ours} && 4.70&7.87\\
\bottomrule
\end{tabular}
\caption{Average content and style loss for the styled images in fig.~\ref{f:comaprison}.  Lower values are better.}
\label{t:numeric_comp}
\end{table}


\textbf{Note:} Gatys approach~\citep{Gatys_16a} should achieve a balanced content and style score similar to ours, but as mentioned in \citet{Sheng_18a} (also shown in third and fourth row in fig.~\ref{f:comaprison}) \citet{Gatys_16a} suffers from getting stuck at a bad local minimum. This results in higher style loss as shown in table~\ref{t:numeric_comp}.



\section{Ablation study}

\subsection{Importance of rigid alignment}
\begin{figure}[t!]
 \begin{tabular}{@{}c@{\hspace{0.4ex}}c@{\hspace{1ex}}c@{\hspace{0.4ex}}c@{\hspace{0.3ex}}c@{}}
 {Content}&{Style} &{Only MM}&{Only RA}&{MM+RA}\\
\includegraphics*[width = 0.19\columnwidth]{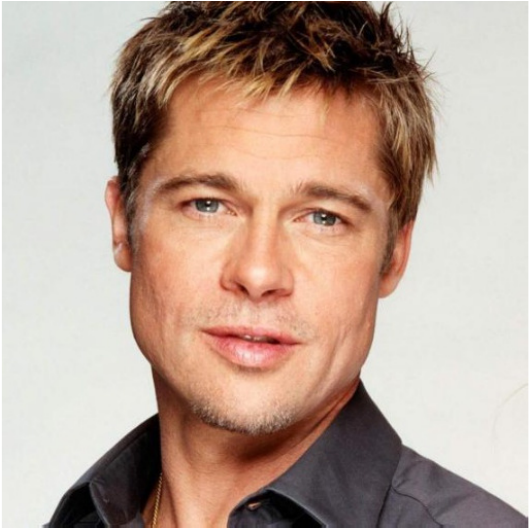} &
\includegraphics*[width = 0.19\columnwidth]{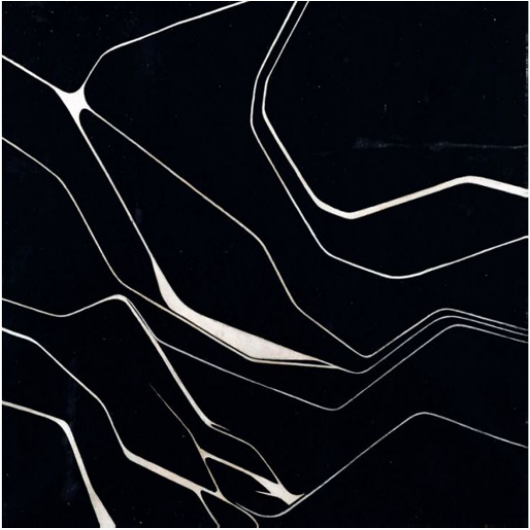}&
\includegraphics*[width=0.19\columnwidth]{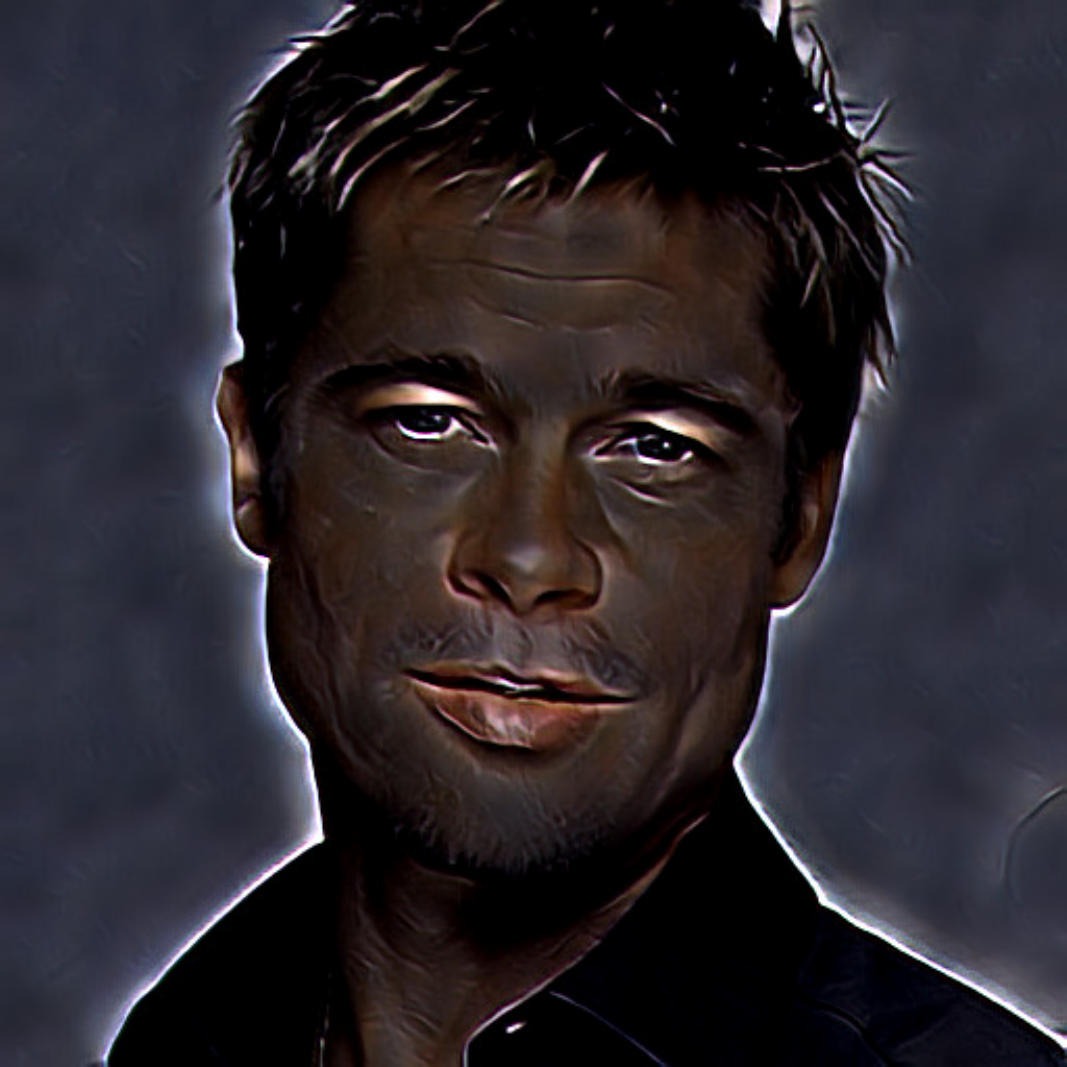}&
\includegraphics*[width=0.19\columnwidth]{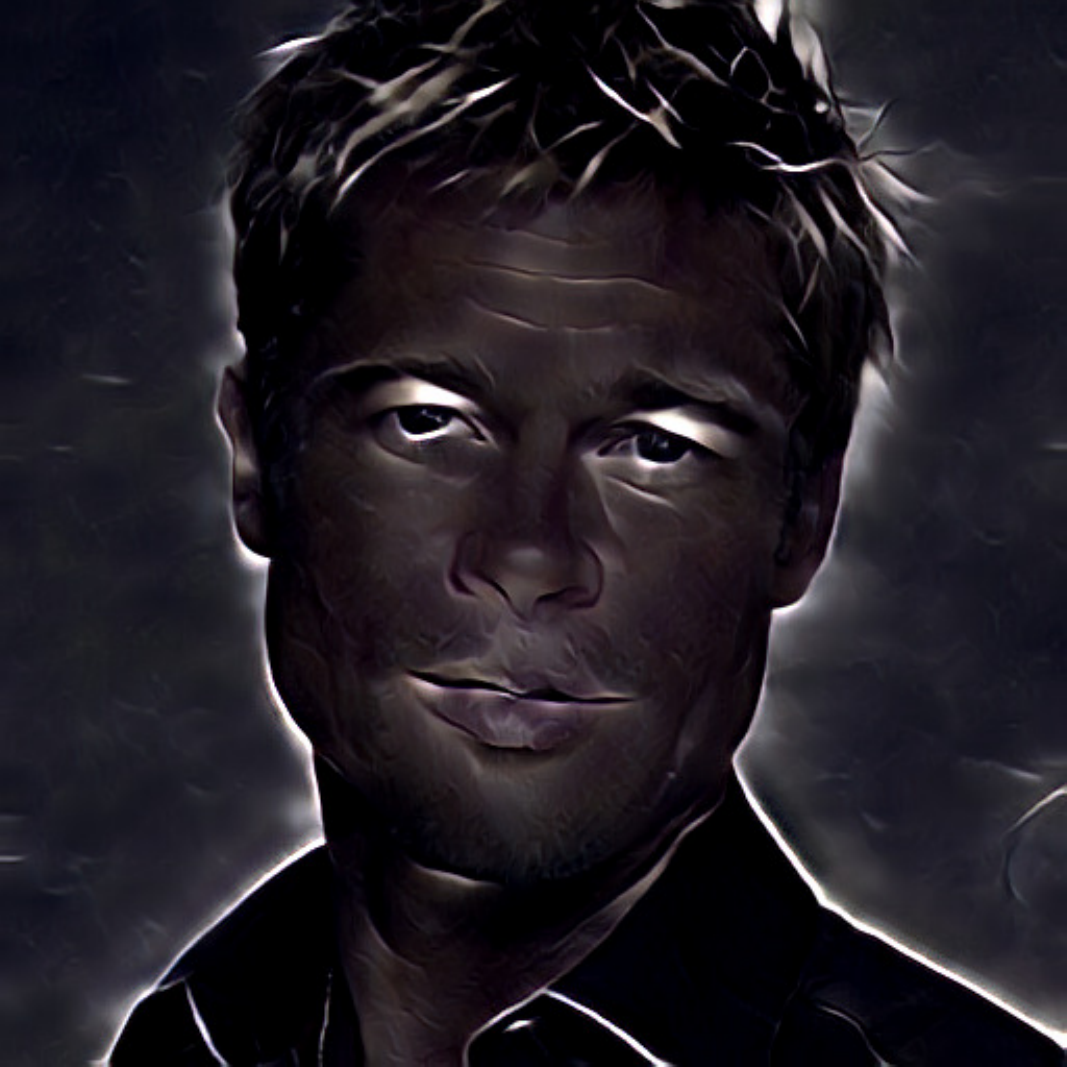}&
\includegraphics*[width=0.19\columnwidth]{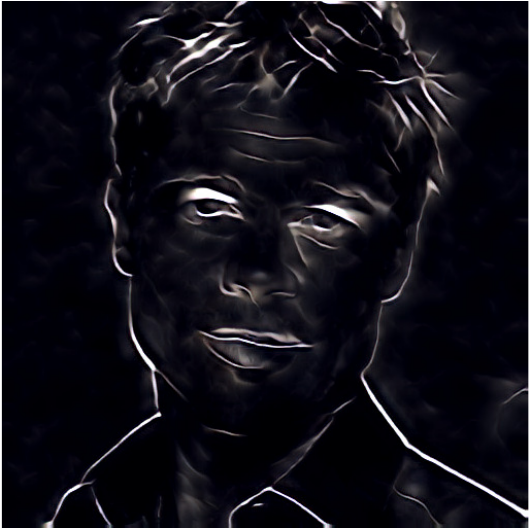}
\end{tabular}
\caption{Style transfer with only moment matching (third column), only rigid alignment (fourth column), and the proposed method (fifth column).}
\label{f:Ab1}
\end{figure}

As described above, our method achieves style transfer by first matching the channel-wise statistics of content features to those of style features and then align style features to content features by rigid alignment. To examine the effect of rigid alignment, we perform the following experiment. We perform style transfer similar to the pipeline described in the section~\ref{s:mutlistyle}, but we remove rigid-alignment (RA) in the deepest layer (relu4\_1). As shown the fig.~\ref{f:Ab1}, moment matching (MM) only transfers low-level style details (in this case, color) while keeping the content structure intact. On the other hand, if we use only rigid alignment, it mostly transfers global style patterns (white strokes around the hair, second column). Finally, when both are used together (proposed method), the resulting image has both global and local style patterns; and thus achieves better-styled results without introducing content distortion. 

\subsection{Cost of preserving content with higher content weight}
\begin{figure}[t]
 \centering
 \begin{tabular}{@{}c@{\hspace{2ex}}c@{\hspace{0.4ex}}c@{\hspace{0.3ex}}c@{\hspace{0.3ex}}c@{}}
 {Content}&&$\alpha =0.9$&$\alpha =0.5$&$\alpha =0.1$\\
 \includegraphics*[width = 0.24\columnwidth]{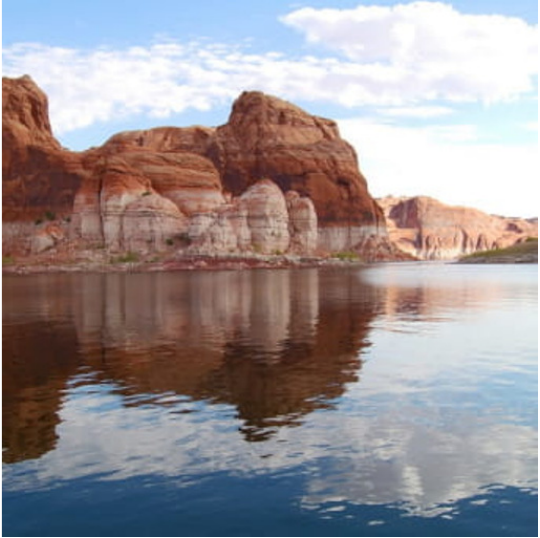} &
 \rotatebox{90}{\hspace{10ex}{WCT}}&
\includegraphics*[width=0.24\columnwidth]{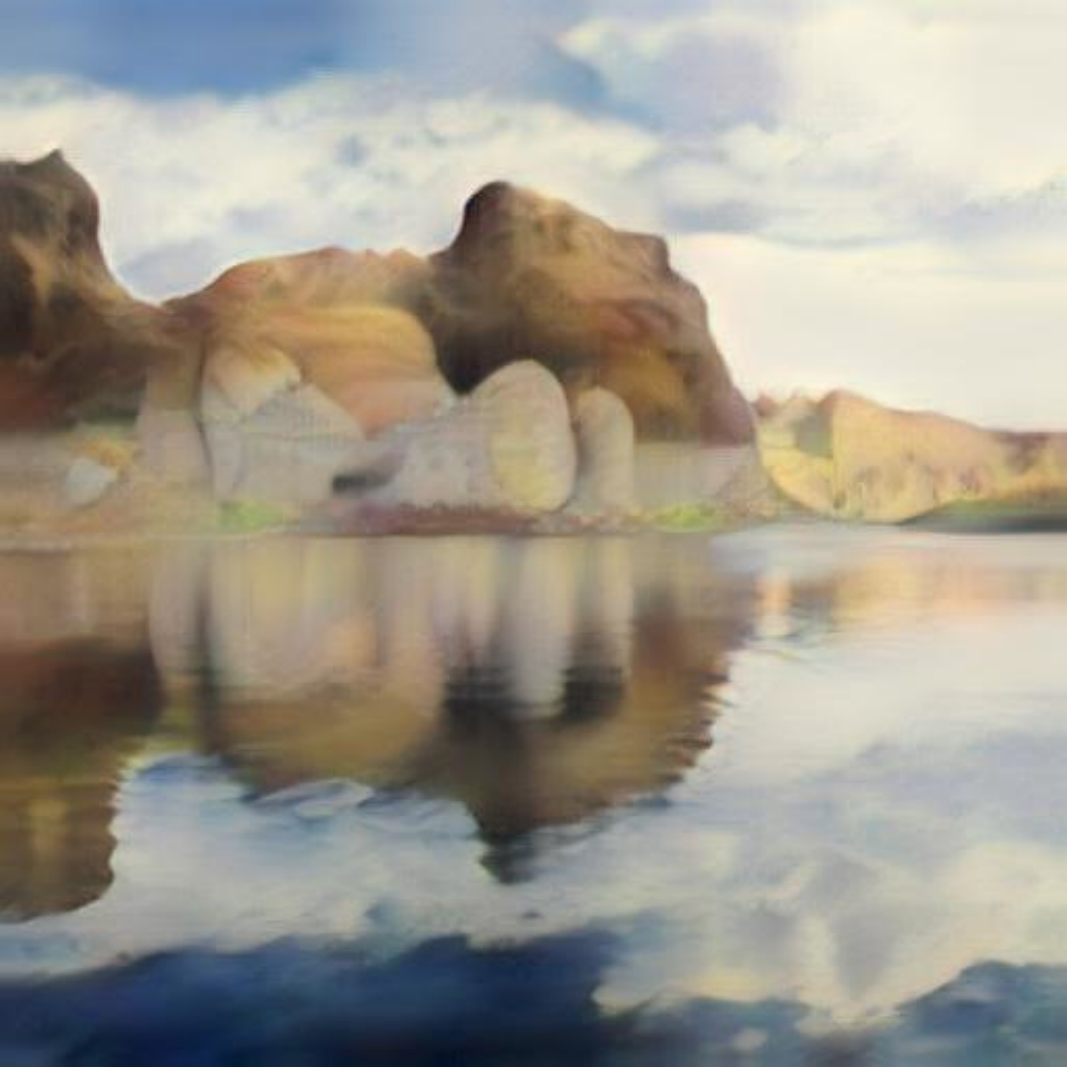}&
\includegraphics*[width=0.24\columnwidth]{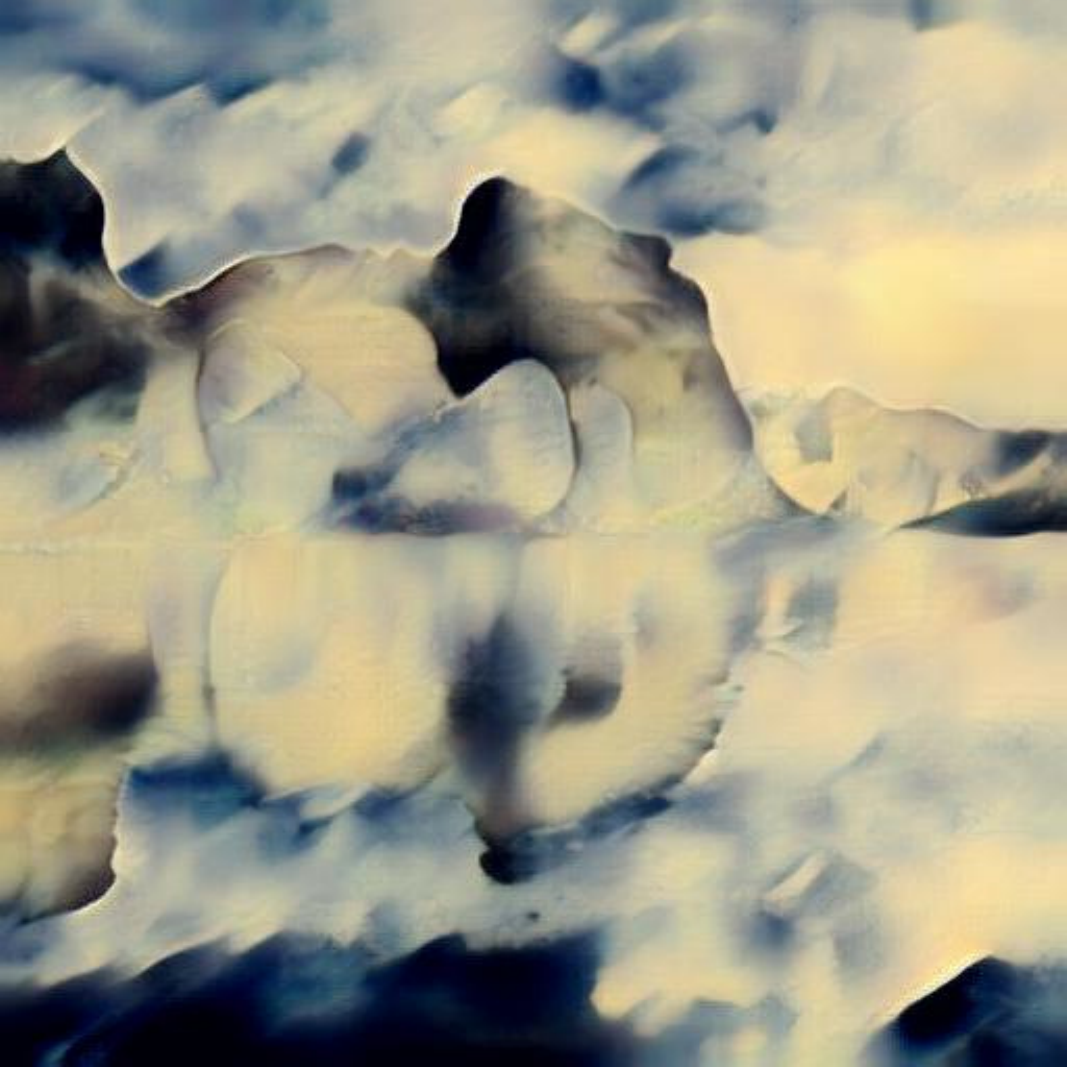}&
\includegraphics*[width=0.24\columnwidth]{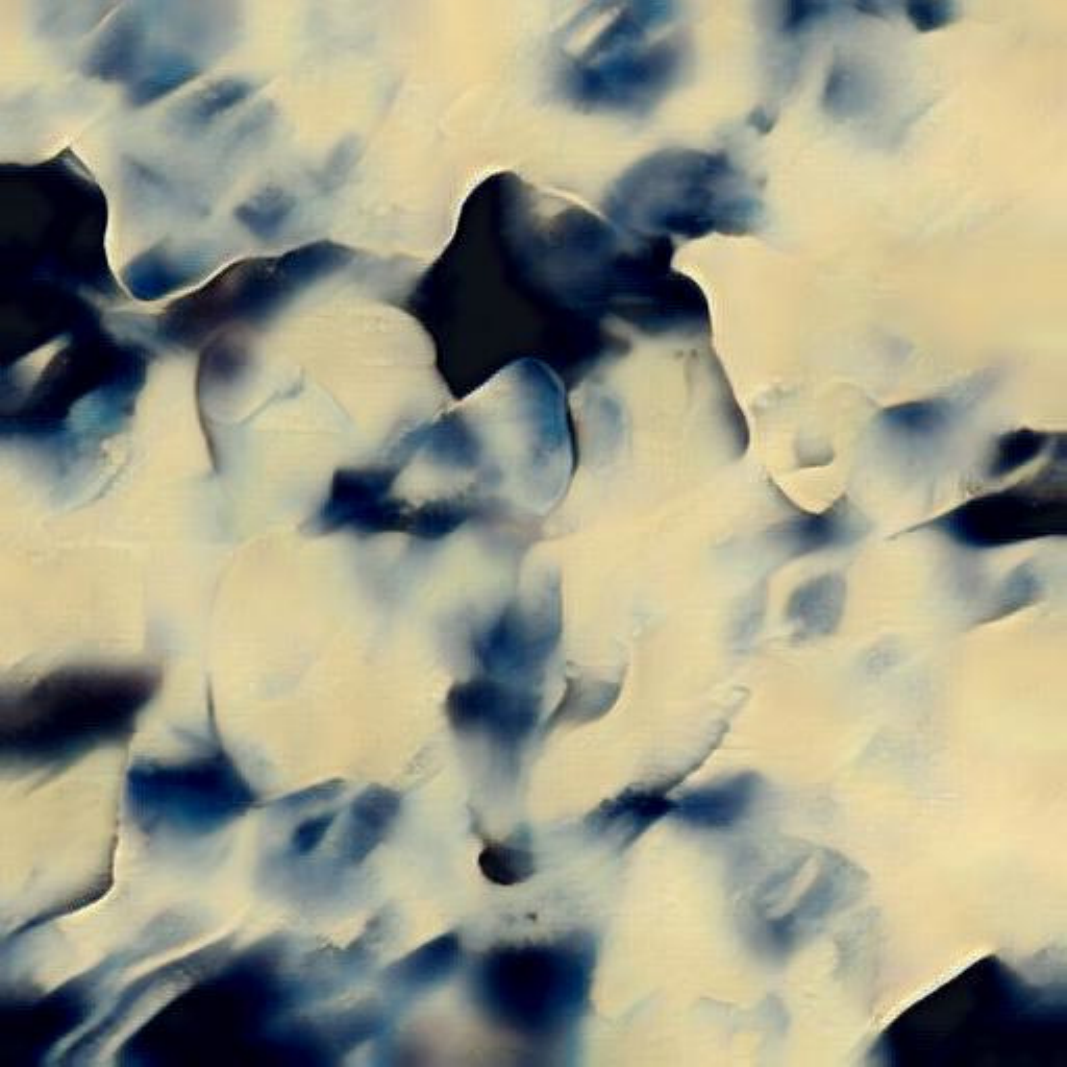}\\
{Style} &&&&\\
\includegraphics*[width = 0.24\columnwidth]{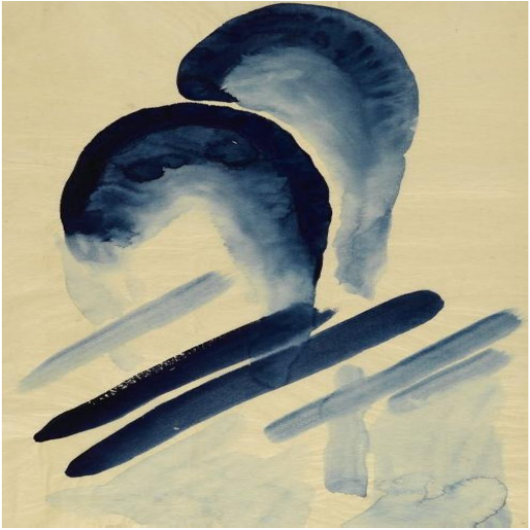}&
 \rotatebox{90}{\hspace{10ex}{Ours}}&
\includegraphics*[width=0.24\columnwidth]{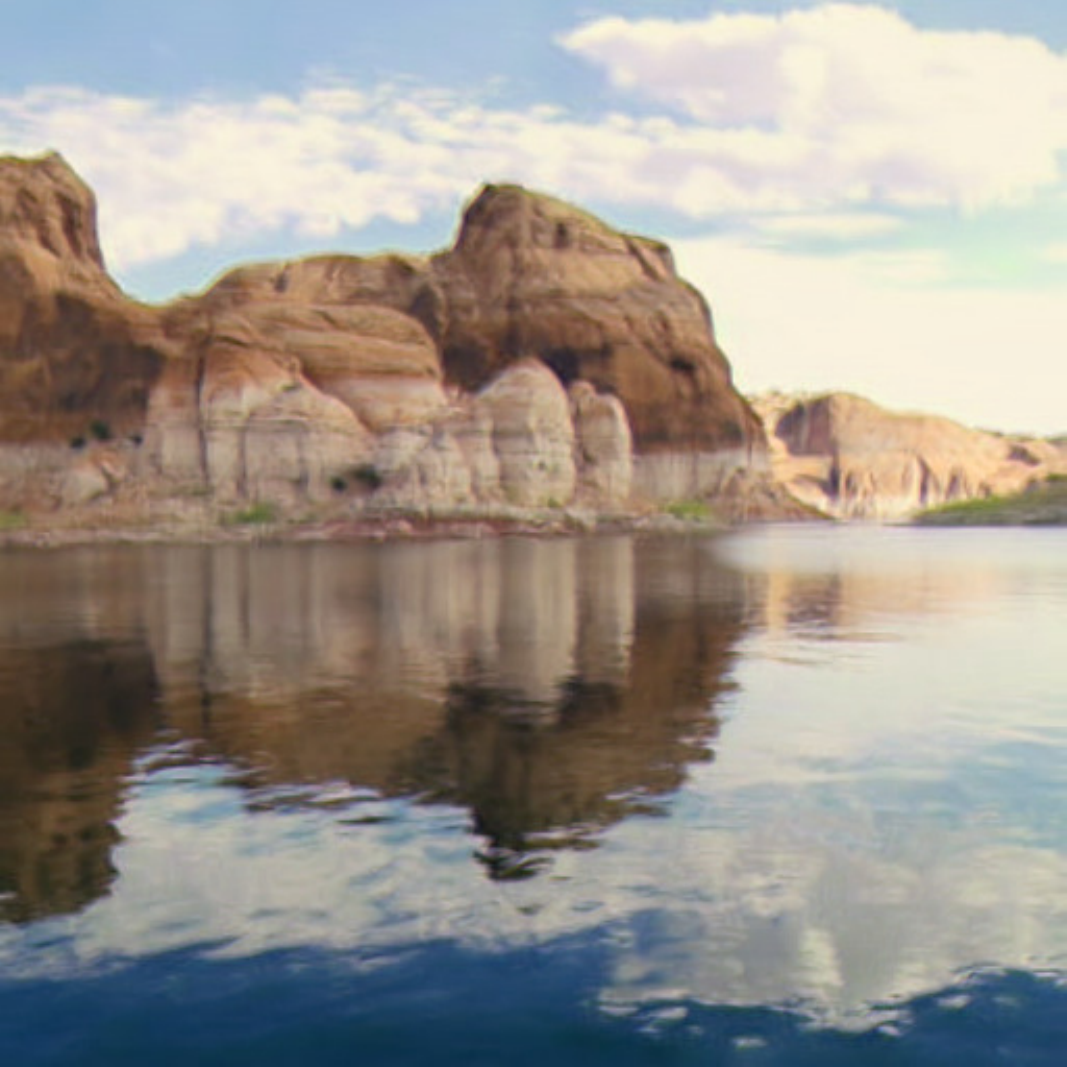}&
\includegraphics*[width=0.24\columnwidth]{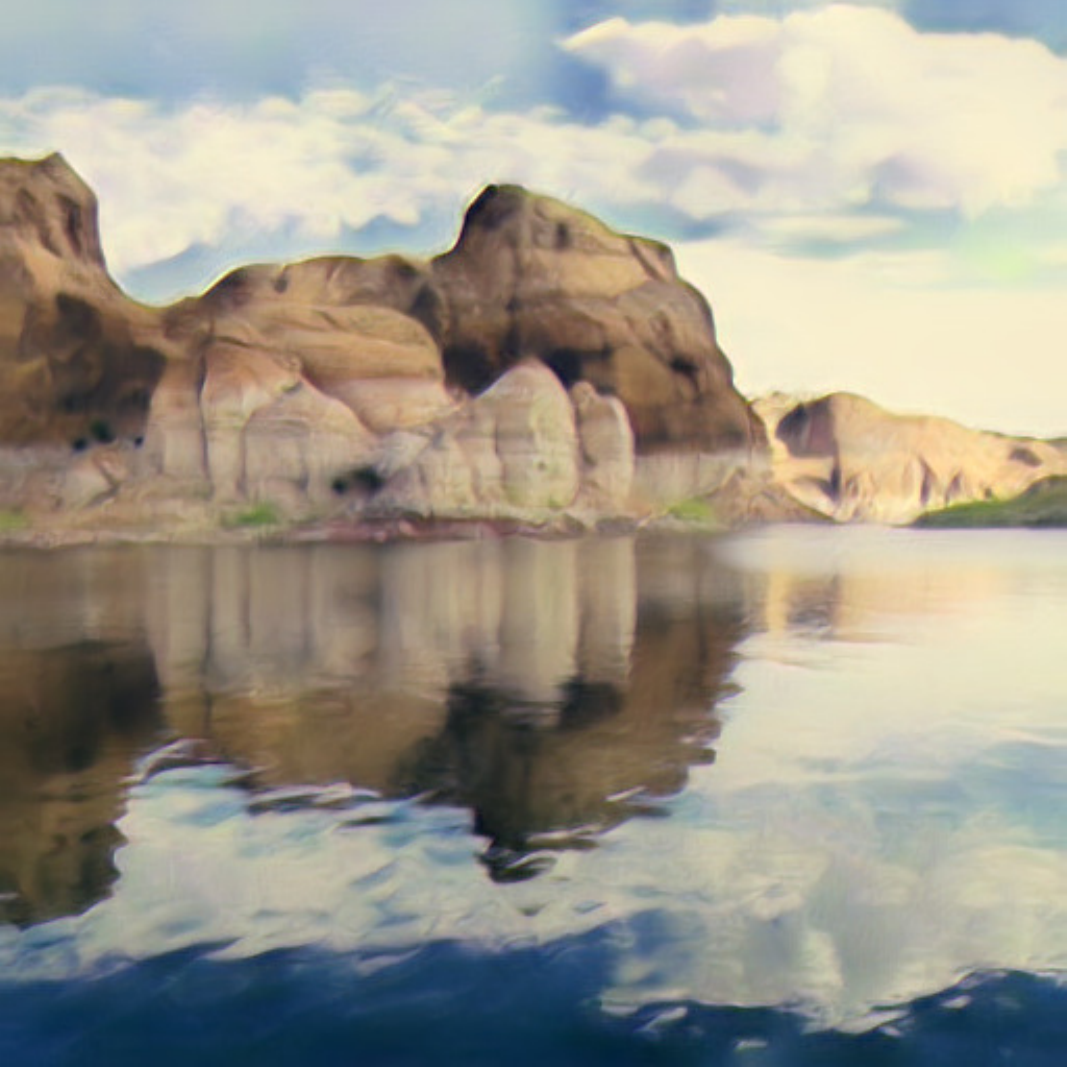}&
\includegraphics*[width=0.24\columnwidth]{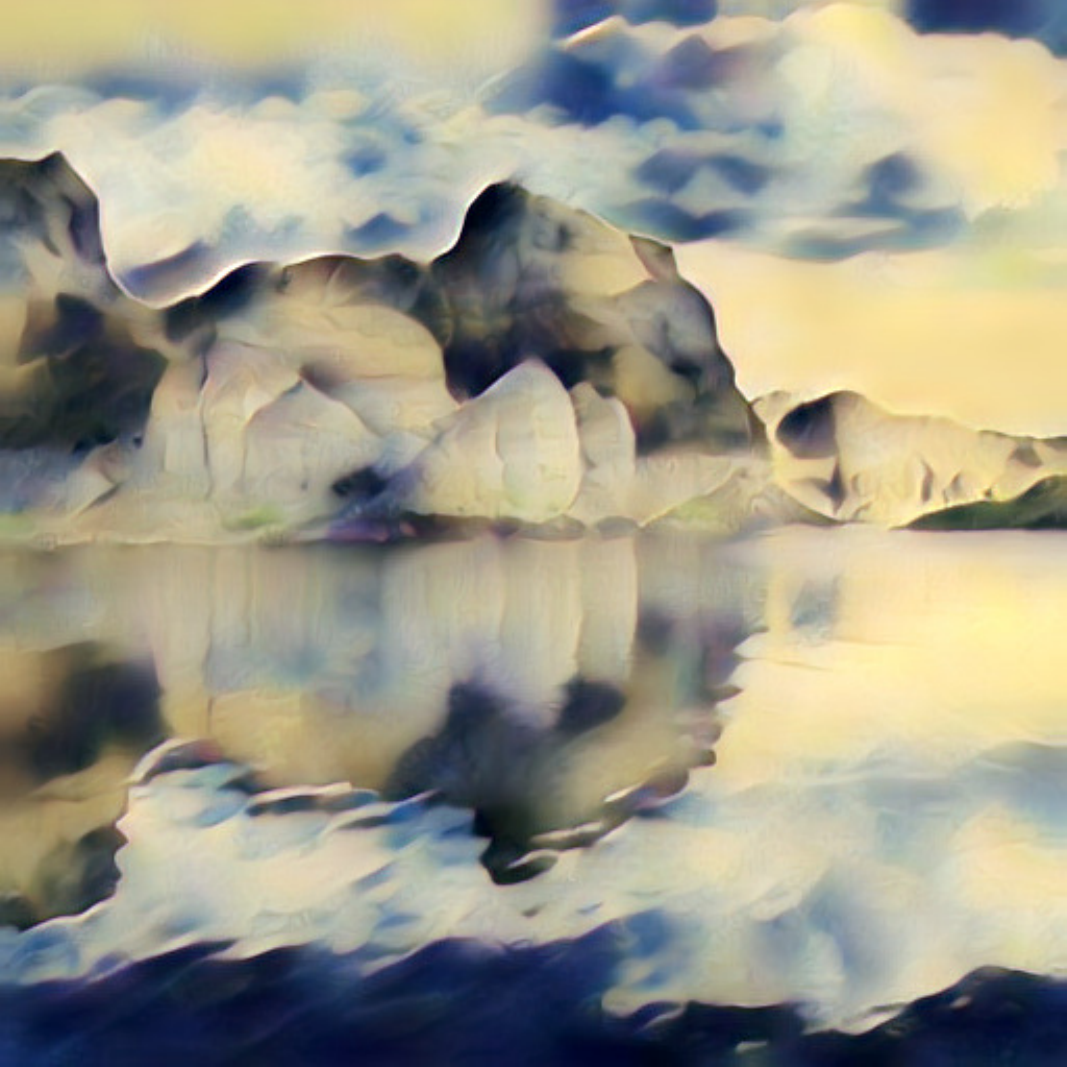}
\end{tabular}
\caption{\emph{Top column 2-4:} style transfer using WCT. \emph{Bottom column 2-4:} style transfer with the proposed approach.}
\label{f:Ab2}
\end{figure}

It can be argued that the content structure can be preserved by adjusting the content weight ($\alpha$). However, having more content weight comes at the cost of ineffective style transfer. In fig.~\ref{f:Ab2}, we show one such example, where we compare this trade-off in case of existing work (we use WCT as an example) and the proposed method. In case of previous works, to preserve the content structure, higher content weight is required; but this results in the insufficient transfer of style patterns (first column). On the other hand, for sufficient transfer of style patterns, content weight needs to be reduced; but this creates distorted content in the styled image (third column in the last row). Our method solves this problem effectively; it not only transfers sufficient style patterns, but also preserves the content structure (last column).  

\begin{figure*}[t!]
 \centering
\begin{tabular}{@{}c@{}}
 {\hspace{0.00\linewidth}style \hspace{0.83\linewidth}content} \\
 { $ \hspace{-0.02\linewidth}\xrightarrow{\hspace{0.3\linewidth} {\normalsize \alpha} \hspace{0.3\linewidth}} $}\\
\includegraphics*[width=1\linewidth]{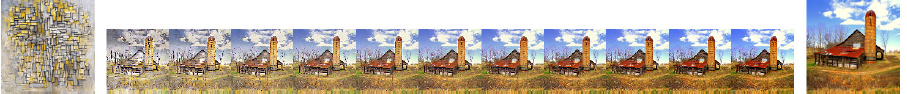}
\end{tabular}
\caption{Trade-off between content and style during style transfer. Value of $\alpha$  is increasing from $0$ to $1$ with an increment of $0.1$ from left to right. }
\label{f:tradeoff}
\end{figure*}

\begin{figure*}[t!]
 \centering
\begin{tabular}{@{}c@{}c@{\hspace{1ex}}|c@{} c@{} c@{}}
{content}&&{style}&{\hspace{0.73\linewidth}style}\\
{image}&&{image 1}&{\hspace{0.73\linewidth}image 2}\\
&&&{ $\xrightarrow{\hspace{0.3\columnwidth}{ \normalsize \beta} \hspace{0.3\columnwidth}} \hspace{0.08\linewidth}$}&\\
\includegraphics*[width=0.08\linewidth]{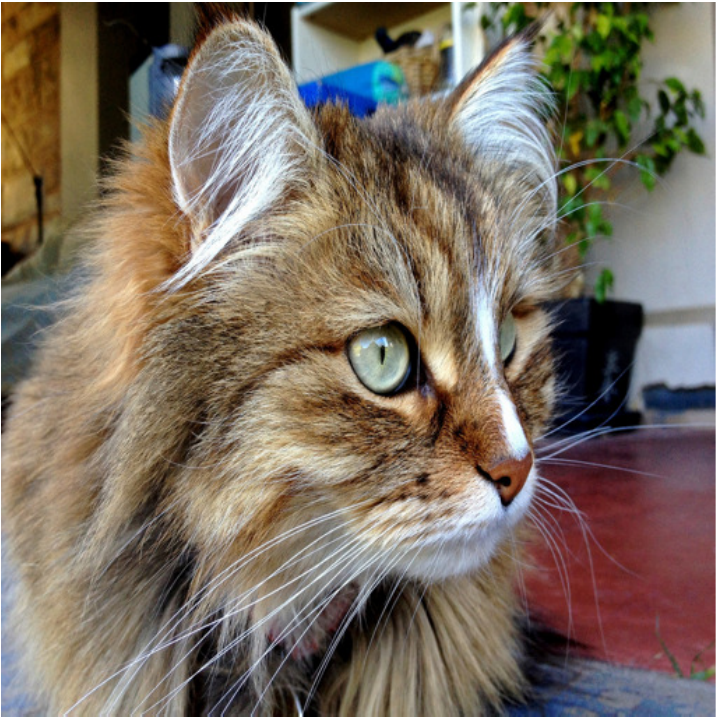}&&
\multicolumn{3}{c}{\includegraphics*[width=0.9\linewidth]{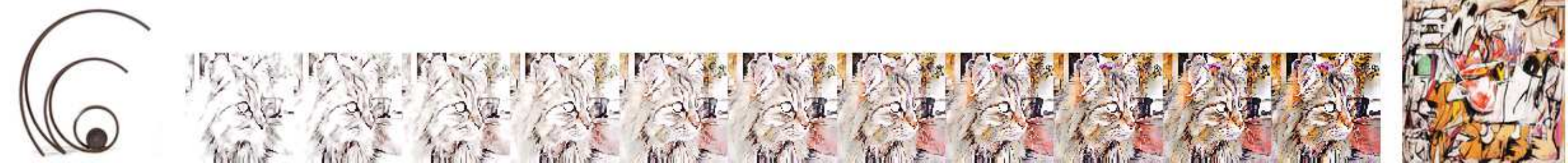}}
\end{tabular}

\caption{Interpolation between styles. Value of $\beta$  is increasing from $0$ to $1$ with an increment of $0.1$ from left to right. }
\label{f:interpolate}
\end{figure*}

\begin{figure*}[t!]
 \centering
 \begin{tabular}{@{}c@{}}
 \includegraphics*[width=1\linewidth]{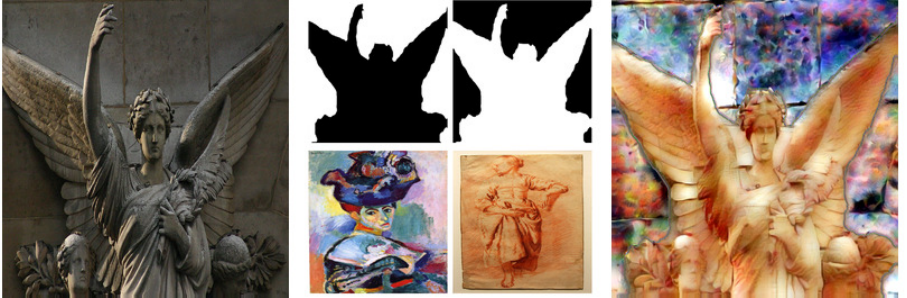}
 \end{tabular}
\caption{Spatial control in style transfer. \emph{Middle column:} In the top row are the binary masks, and corresponding styles are in the bottom row. }
\label{f:mask}
\end{figure*}

\section{User control}
Like other arbitrary style transfer methods, our approach is also flexible to accommodate different user controls such as the trade-off between style and content, style interpolation, and spatial control during style transfer. 

Since our method applies transformation in the feature-space independent of the network, we can achieve trade-off between style and content as follows: 
\begin{align}
\z = \alpha \z_c + (1-\alpha) \z_{sc} .
\end{align}
Here, $\z_{sc}$ is the transformed feature from eq.~\eqref{e:proc}, $\z_c$ is content feature and $\alpha $ is the trade off parameter. Fig.~\ref{f:tradeoff} shows one such example of content-style trade-off.

Fig.~\ref{f:interpolate} shows an instance of linear interpolation between two styles created by proposed approach. This is done by adjusting the weight parameter ($\beta$) between transformation outputs ($\calT(\z_c,\z_s)$) as follows:
\begin{align}
\z = \alpha \z_c + (1-\alpha) (\beta \calT(\z_{c} ,\z_{s1}) + (1-\beta) \calT (\z_c,\z_{s2}) ) .
\end{align}
Spatial control is needed to apply different styles to different parts of the content image. A set of masks $\M$ are additionally required to control the regions of correspondence between style and content. By replacing the content feature $\z_c$ in section~4 of the main paper with $\M \odot \z_c$, where $\odot$ is a simple mask-out operation, we can stylize the specified region only, as shown in figure~\ref{f:mask}.

\section{Conclusion}

In this work, we propose an effective arbitrary style transfer approach that does not require learning for every individual style. By applying rigid alignment to style features with respect to content features, we solve the problem of content distortion without sacrificing style patterns in the styled image. Our method can seamlessly adapt the existing multi-layer stylization pipeline and capture style information from those layers too.  
Our method can also seamlessly perform video stylization, merely by per-frame style transfer. Experimental results demonstrate that the proposed algorithm achieves favorable performance against the state-of-the-art methods in arbitrary style transfer. As a further direction, one may replace multiple autoencoders for multi-level style transfer by training an hourglass architecture similar to Avatar-Net for better efficiency.  

\vspace{10ex}

\appendix 

\section{More styled Results}

\begin{figure*}[h!]
 \centering
 \begin{tabular}{@{\hspace{5ex}}c@{\hspace{9ex}}  c@{\hspace{5ex}}c@{\hspace{5ex}} c@{\hspace{5ex}}  c@{\hspace{5ex}} c@{\hspace{4ex}}  c@{\hspace{2ex}}  c@{}}
 {content}& {style} && {Gatys[\citenum{Gatys_16a}] }& {AdaIN [\citenum{HuangBelong17a}]} & {WCT [\citenum{Li_17d}]} & {Avatar [\citenum{Sheng_18a}]}& {Ours}\\
\multicolumn{8}{c}{\includegraphics*[width = 1\linewidth]{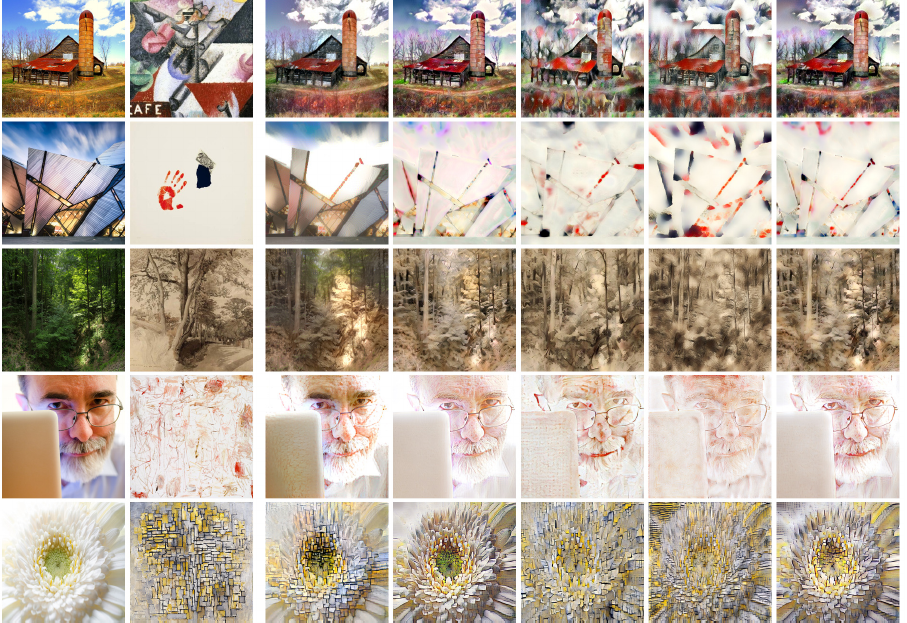}}

\end{tabular}
\end{figure*}

\begin{figure*}[ht!]
 \centering
 \begin{tabular}{@{\hspace{5ex}}c@{\hspace{9ex}}  c@{\hspace{5ex}}c@{\hspace{5ex}} c@{\hspace{5ex}}  c@{\hspace{5ex}} c@{\hspace{4ex}}  c@{\hspace{2ex}}  c@{}}
 {content}& {style} && {Gatys[\citenum{Gatys_16a}] }& {AdaIN [\citenum{HuangBelong17a}]} & {WCT [\citenum{Li_17d}]} & {Avatar [\citenum{Sheng_18a}]}& {Ours}\\
\multicolumn{8}{c}{\includegraphics*[width = 1\linewidth]{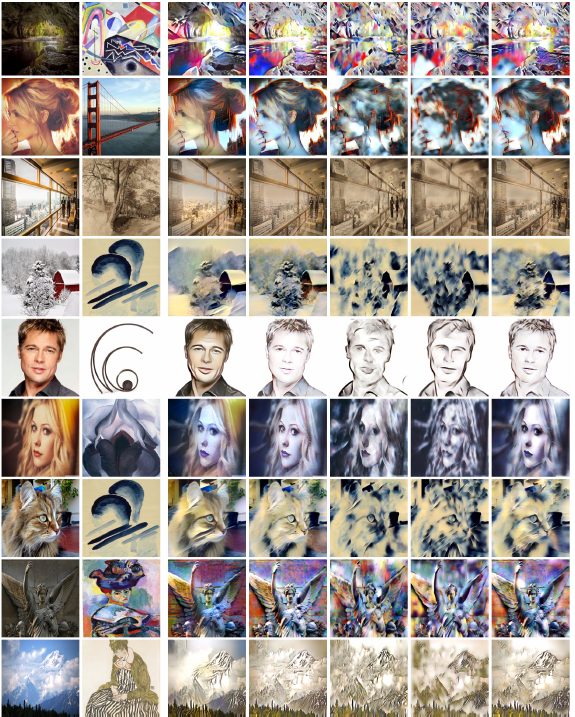}}

\end{tabular}
\end{figure*}

\begin{figure*}[ht!]
 \centering
 \begin{tabular}{@{\hspace{5ex}}c@{\hspace{9ex}}  c@{\hspace{5ex}}c@{\hspace{5ex}} c@{\hspace{5ex}}  c@{\hspace{5ex}} c@{\hspace{4ex}}  c@{\hspace{2ex}}  c@{}}
 {content}& {style} && {Gatys[\citenum{Gatys_16a}] }& {AdaIN [\citenum{HuangBelong17a}]} & {WCT [\citenum{Li_17d}]} & {Avatar [\citenum{Sheng_18a}]}& {Ours}\\
\multicolumn{8}{c}{\includegraphics*[width = 1\linewidth]{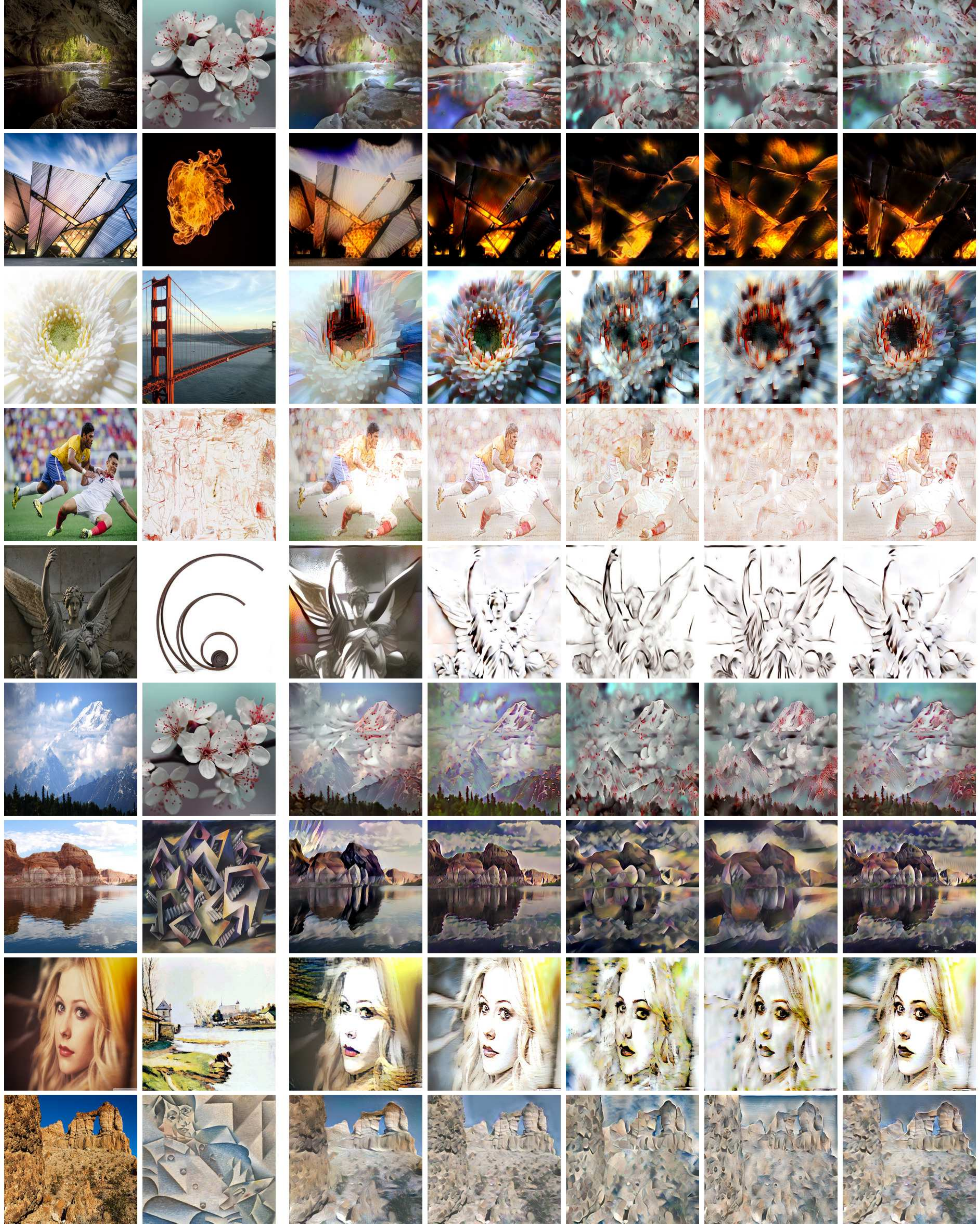}}

\end{tabular}
\end{figure*}

\clearpage
\bibliographystyle{abbrvnat}


\end{document}